\documentclass[11pt,a4paper]{article}
\usepackage[hyperref]{emnlp-ijcnlp-2019}
\usepackage{times}
\usepackage{latexsym}

\usepackage{url}

\aclfinalcopy 

\usepackage[pdftex]{graphicx}
\usepackage{CJKutf8}
\usepackage{url}
\usepackage{CJKutf8}

\title{Corpus Augmentation by Sentence Segmentation \\
 for Low-Resource Neural Machine Translation}

\author{Jinyi Zhang \\
  Graduate School of Engineering \\
  Gifu University \\
  Gifu, Japan \\
  {\tt zhang-jinyi@outlook.com} \\\And
  Tadahiro Matsumoto \\
  Department of Electrical, \\Electronic and Computer Engineering \\
  Gifu University \\
  Gifu, Japan \\
  {\tt tad@gifu-u.ac.jp} \\}

\date{}
\begin{document}
\maketitle
\begin{abstract}
Neural Machine Translation (NMT) has been proven to achieve impressive results. The NMT system translation results depend strongly on the size and quality of parallel corpora. Nevertheless, for many language pairs, no rich-resource parallel corpora exist. As described in this paper, we propose a corpus augmentation method by segmenting long sentences in a corpus using back-translation and generating pseudo-parallel sentence pairs. The experiment results of the Japanese-Chinese and Chinese-Japanese translation with Japanese-Chinese scientific paper excerpt corpus (ASPEC-JC) show that the method improves translation performance.
\end{abstract}

%

\section{Introduction}
\label{section:intro}
Neural Machine Translation (NMT) has produced remarkable results with large-scale parallel corpus. However, for low-resource languages for domain-defined translation tasks, the parallel corpus scale is small. Accordingly, the translation performance is reduced considerably \cite{KoehnK17}. Therefore, the study of NMT under conditions of low-resource language corpora has high practical value.

As described in this paper, 
we propose a corpus augmentation method by segmenting long sentences into partial sentences of the corpus using back-translation and generating pseudo-parallel sentence pairs. The larger corpus can improve translation performance, in experiments on the Japanese--Chinese scientific paper excerpt corpus (ASPEC-JC) as the low-resource corpus, the translation results over the baseline both have better translation performance in Japanese-Chinese and Chinese-Japanese directions, respectively. 

The main contributions of this paper are the following.
We demonstrate the ability to improve the translation performance of NMT systems by mixing generated pseudo-parallel sentence pairs into training data with no monolingual data, and without changing the neural network architecture. This capability makes our approach applicable to different NMT architectures.

\section{Related Work}

Expanding the number of parallel corpora is an effective means of improving the translation quality for NMT in low-resource languages. 
The parallel corpus can be constructed quickly using back-translation with monolingual target data \cite{Sennrich2016b}. One study reported by \citet{Sennrich2017} also showed that even simply duplicating the monolingual target data and using them as the source data was sufficient to realize some benefits. Moreover, a pseudo-parallel corpus can be constructed using the copy method, i.e., the target language sentences are copied as the corresponding source language sentences \cite{Currey2017}, which illustrates that even poor translations can be beneficial. Data augmentation for low-frequency words has also been proven an
effective method \cite{Fadaee2017}. 

For back-translation method, \citet{Gwin} implemented their NMT system with iteratively applying back-translation. \citet{Lample} explored the use of generated back-translated data, aided by denoising with a language model trained on the target side.
Translation performance can also be improved by iterative back-translation in both
high-resource and low-resource scenarios \cite{invest}. A more refined idea of back-translation is the dual learning approach of \citet{He2016}, which
integrates training on parallel data and training on
monolingual data via round-tripping.

\section{NMT and ASPEC-JC Corpus}
\label{section:NMT-JC}
For this research, we follow the NMT architecture by \citet{Luong2015}, which implements as a global attentional encoder--decoder neural
network with Long Short-Term Memory (LSTM). 
We simply use it at the character level, because the translation results have better performance than the word-level between Japanese and Chinese. 
However, it is noteworthy that our proposed method is not specific to this architecture.

We conducted experiments with the ASPEC-JC corpus, which was constructed by manually translating Japanese scientific papers into Chinese \cite{NAKAZAWA16.621}. 
ASPEC-JC comprises four parts: training data (672,315 sentence pairs), development data (2,090 sentence pairs), development-test data (2,148 sentence pairs) and test data (2,107 sentence pairs) on the assumption that they would be used for machine translation research.

We chose ASPEC-JC as the low-resource corpus compared with other language pairs such as English-French, which usually comprises millions of parallel sentences. ASPEC-JC corpus only has about 672k sentences. We randomly extracted 300k sentence pairs from the training data for experiments.

\section{Corpus Augmentation by Long Sentence Segmentation}
\label{section:aug}
\citet{Sennrich2016b} proposed a method to extend parallel corpora
by back-translating target language sentences in monolingual corpora
to obtain pseudo-source sentences; 
the pseudo-source sentences together with the
original target sentences are then added to the parallel corpus.

Our method expands the existing parallel corpus
with itself, not with any monolingual data, not like some back-translation methods with monolingual data \cite{Sennrich2016b} \cite{Currey2017} \cite{Fadaee2017}. Moreover, our method could be combined with other corpus augmentation methods. 
Our augmentation process includes the
following phases: 1) splitting `\textit{long}' parallel sentence pairs
of the corpus into parallel partial sentence pairs,
2) back-translating the target partial sentences,
and 3) constructing parallel sentence pairs by combining the source
and the back-translated target partial sentences. To be precise, a
`\textit{long}' sentence above means a sentence that contains more than one punctuation marks.

\subsection{Generating bilingual partial sentences}
\label{subsection:generate}

The following procedure generates parallel partial sentence pairs from
\textit{long} parallel sentence pairs.

\begin{figure*}[bth]
\begin{center}
\includegraphics[width=150mm]{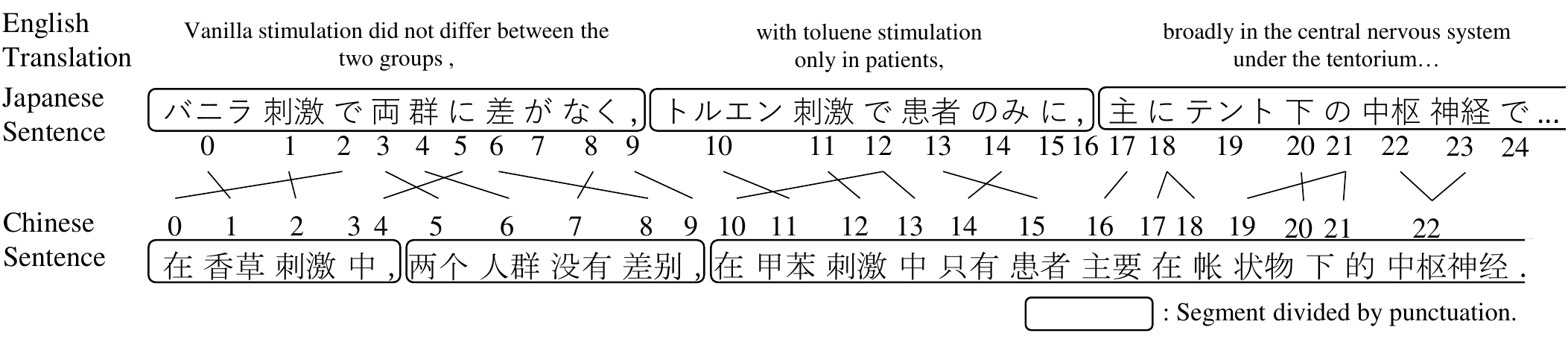}
\caption{Example of word alignment information and sentence segments.}
\label{fig:word-alignment}
\end{center}
\end{figure*}

\begin{figure*}[bth]
\begin{center}
\includegraphics[width=150mm]{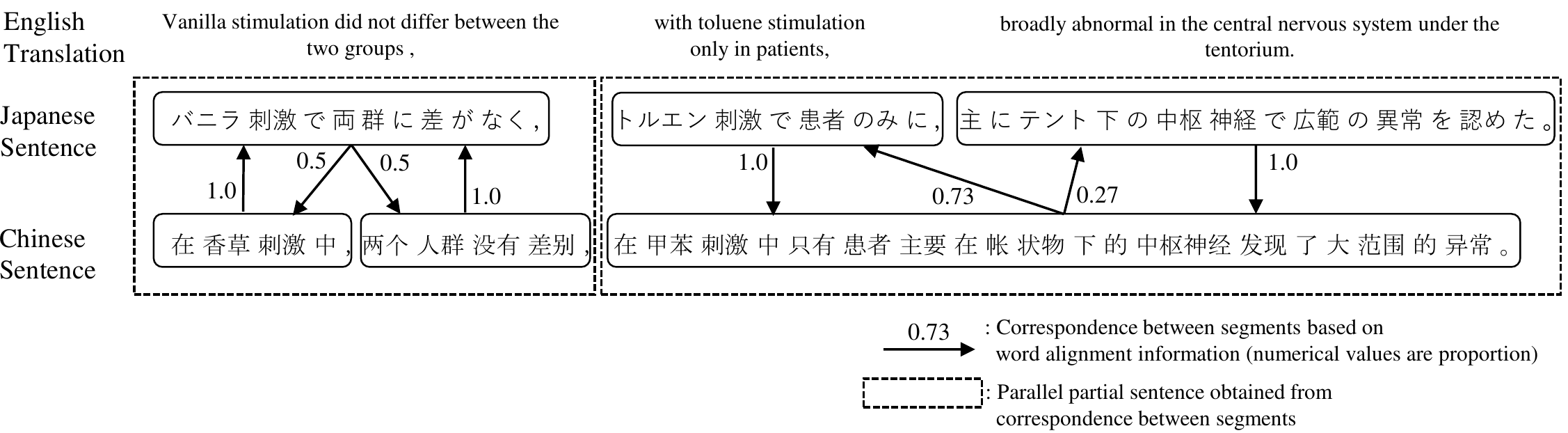}
\caption{Examples of generated parallel partial sentences.}
\label{fig:alignment}
\end{center}
\end{figure*}

\begin{enumerate}
\item Obtain the word alignment information from tokenized
    Japanese-Chinese parallel sentences.
\item Split the \textit{long} parallel sentences into segments at the punctuation symbols,
 such as ``,'', ``;'', ``:''.
Figure \ref{fig:word-alignment} presents an example of the word alignment
information and the segments of a sentence pair.

\item Obtain source-target segment alignments: For each source segment $\mathrm{s\mathchar`-seg}_i$ and target segment $\mathrm{t\mathchar`-seg}_j$, count the words in $\mathrm{s\mathchar`-seg}_i$ that correspond to the words in $\mathrm{t\mathchar`-seg}_j$ according to the word alignment information. The numerical values on the arrows in Figure \ref{fig:word-alignment} represent the rate of the correspondence relation between the segments. We infer that $\mathrm{s\mathchar`-seg}_i$ corresponds to $\mathrm{t\mathchar`-seg}_j$ if the rate is greater than or equal to a threshold value $\theta$. In this research, we set $\theta=0.5$.

\item Obtain target-source segment alignments: According to the procedure in 3.
\item Concatenate multiple segments to form a one-to-one relation if there is a one-to-many or many-to-many relation between the segments.
\end{enumerate}

In Figure \ref{fig:alignment}, each sentence is
divided into three segments. Thereby, two parallel partial sentences are generated.

\begin{table*}[bt]
\centering
\caption{Experiment results of 300k training data. Translation directions are designated as Japanese$\rightarrow$Chinese (JC, J$\rightarrow$C) and
Chinese$\rightarrow$Japanese (CJ, C$\rightarrow$J) in the table. }
\label{tab:results-300k}
\begin{tabular}{|c|c|c|c|c|c|c}
\hline

  \multicolumn{1}{|c|}{\raisebox{-2ex}{\bf Method}} &
  \multicolumn{1}{c|}{\raisebox{-1ex}{\bf \#}} &
  \multicolumn{1}{c|}{\raisebox{-1ex}{\bf \# back-}} &
  \multicolumn{1}{c|}{\raisebox{-0.8ex}{\bf J$\rightarrow$C}} &
  \multicolumn{1}{c|}{\raisebox{-0.8ex}{\bf C$\rightarrow$J}} \\ 
  
  \multicolumn{1}{|c|}{}&
  \multicolumn{1}{c|} {\raisebox{0.5ex}{\bf sentences}} & 
  \multicolumn{1}{c|} {\raisebox{0.5ex}{\bf translated}} & 
  \multicolumn{1}{c|}{\raisebox{0ex}{\bf BLEU (\%)}} &
  \multicolumn{1}{c|}{\raisebox{0ex}{\bf BLEU (\%)}} \\ \hline
  
\multicolumn{1}{|c|}{Baseline} &   \multicolumn{1}{c|}{300k(JC),300k(CJ)} & \multicolumn{1}{c|}{0} &\multicolumn{1}{c|}{38.7} & \multicolumn{1}{c|}{37.9} \\ 
\multicolumn{1}{|c|}{Copied}   & \multicolumn{1}{c|}{952k(JC),952k(CJ)} & \multicolumn{1}{c|}{0} &\multicolumn{1}{c|}{39.2 (+0.5)} & \multicolumn{1}{c|}{39.8 (+1.9)} \\ 
\multicolumn{1}{|c|}{Partial}  & \multicolumn{1}{c|}{984k(JC),984k(CJ)} &\multicolumn{1}{c|}{0} &\multicolumn{1}{c|}{39.2 (+0.5)} & \multicolumn{1}{c|}{39.2 (+1.3)} \\ 
\multicolumn{1}{|c|}{Back-translation}  & \multicolumn{1}{c|}{518k(JC),518k(CJ)} &\multicolumn{1}{c|}{218k(JC),218k(CJ)} &\multicolumn{1}{c|}{39.4 (+0.7)} & \multicolumn{1}{c|}{39.4 (+1.5)} \\ 
\multicolumn{1}{|c|}{Proposed}  & \multicolumn{1}{c|}{952k(JC),952k(CJ)} & \multicolumn{1}{c|}{218k(JC),218k(CJ)} &\multicolumn{1}{c|}{\bf 39.5 (+0.8)} & \multicolumn{1}{c|}{\bf 40.1 (+2.2)} \\ \hline

\end{tabular}

\vspace*{0.8ex}
\caption{Experiment results of 300k training data with 372k monolingual data. Translation directions are designated as Japanese$\rightarrow$Chinese (JC, J$\rightarrow$C) and
Chinese$\rightarrow$Japanese (CJ, C$\rightarrow$J) in the table.}
\label{tab:results-mono}
\begin{tabular}{|c|c|c|c|c|c|c}
\hline

  \multicolumn{1}{|c|}{\raisebox{-2ex}{\bf Method}} &
  \multicolumn{1}{c|}{\raisebox{-1ex}{\bf \#}} &
  \multicolumn{1}{c|}{\raisebox{-1ex}{\bf \# back-}} &
  \multicolumn{1}{c|}{\raisebox{-0.8ex}{\bf J$\rightarrow$C}} &
  \multicolumn{1}{c|}{\raisebox{-0.8ex}{\bf C$\rightarrow$J}} \\ 
  
  \multicolumn{1}{|c|}{}&
  \multicolumn{1}{c|} {\raisebox{0.5ex}{\bf sentences}} & 
  \multicolumn{1}{c|} {\raisebox{0.5ex}{\bf translated}} & 
  \multicolumn{1}{c|}{\raisebox{0ex}{\bf BLEU (\%)}} &
  \multicolumn{1}{c|}{\raisebox{0ex}{\bf BLEU (\%)}}  \\ \hline
  
\multicolumn{1}{|c|}{300k+mono} &   \multicolumn{1}{c|}{672k(JC),672k(CJ)} & \multicolumn{1}{c|}{372k(JC),372k(CJ)} &\multicolumn{1}{c|}{39.6} & \multicolumn{1}{c|}{39.9} \\ 

\multicolumn{1}{|c|}{+Proposed}  & \multicolumn{1}{c|}{2,255k(JC),2,200k(CJ)} & \multicolumn{1}{c|}{508k(JC),513k(CJ)} &\multicolumn{1}{c|}{\bf 40.1(+0.5)} & \multicolumn{1}{c|}{\bf 41.1(+1.2)} \\ \hline

\end{tabular}
\end{table*}

\subsection{Corpus augmentation by generated bilingual partial sentences}
\label{subsection:process}
Using the generated parallel partial sentences,
pseudo-parallel sentences are constructed according to the following procedure.

\begin{enumerate}
\item Back-translate the target partial sentences into source language
with a translation model built from parallel data.

\item Create a pseudo-source sentence that is partly different from
the original source sentence by replacing a part of the original sentence with
a partial sentence obtained through back-translation.
For example, if a sentence is divided into three partial sentences, then
three pseudo-source sentences will be created.

\item Copy the target sentences corresponding to the created pseudo-source sentences
to produce pseudo-parallel sentences.

\item Add the generated pseudo-parallel sentences to the original parallel corpus.
\end{enumerate}

\section{Evaluation and Discussion}
\label{section:eva}
\subsection{Experiment settings}

We follow the NMT architecture by
\citet{Luong2015} and implement the NMT architecture using OpenNMT \cite{OpenNMT:Klein}. 
The model has one layer with 512 cells; the embedding size is 512. The parameters are uniformly initialized in ($-0.1, 0.1$), using plain SGD, starting with a learning rate of 1 until epoch 6, and subsequently 0.5 times for each epoch. The max-batch size is 100. The normalized gradient is rescaled whenever its norm exceeds 1. Because of the amounts of training data (300k as the baseline) is relatively small, the dropout probability is set
as 0.5 to avoid overfitting. Decoding is performed by beam search with a beam size of 5. 
We segment the Chinese and Japanese sentences into words using Jieba \footnote{http://github.com/fxsjy/jieba} and Mecab \footnote{http://taku910.github.io/mecab}. We employed fast\_align to obtain word alignment information, which was symmetrized using the included atool command\footnote{http://github.com/clab/fast\_align}. 

The average of BLEU scores from validation perplexity (perplexity with dev data) stopped point to epoch 16 was taken as the evaluation BLEU value. 

\subsection{Experiment results and discussion}

The translation results are presented in Table
\ref{tab:results-300k} (for 300k sentence pairs).  ``Baseline'' is a character-level translation with the 300k original training data. The back-translation models for corpus augmentation are constructed using the 300k original training data of ``Baseline''. 
``Copied'' is the method that adds duplicate copies of both the source and target sides of the training data as the same
times as the proposed method does. The experiment of this method aims to highlight
differences between the generated pseudo-parallel sentences pairs and
unchanged sentences pairs. ``Partial'' is the method that augments the corpus with parallel partial sentences generated by the procedure in Section \ref{subsection:generate}, without back-translating and mixing the partial sentences. The experiment of this method aims to confirm the mixing step (Section \ref{subsection:generate}, step 2) is necessary. This method expands the parallel corpus from 300k sentence pairs to 984k sentence pairs in both directions. 
``Back-translation'' is the back-translation method that back-translates the same data as the proposed method does (218k from original training data). The experiment of this method aims to compare proposed method with the back-translation method (\citet{Sennrich2016b}) on the same back-translated data. 
``\# sentences'' in the tables denotes the size (the number of sentence
pairs) of training data, 
whereas ``\# back-translated'' denotes the number of parallel
sentence pairs used for back-translation processing, i.e., the corpus augmentation, in each method.  


Although the generated pseudo-source sentences have translation errors and unnatural expressions, the BLEU scores were higher than ``Copied'', `Partial'' and ``Back-translation''
in both directions: J$\rightarrow$C and C$\rightarrow$J.
These results demonstrated that the proposed method is effective for extending
the small-scale parallel corpus to improve NMT performance.


The experiments described above prove the effectiveness of the proposed method. Nevertheless, our approach is based on only the original parallel data and does not require any additional monolingual data, unlike back-translation method of \citet{Sennrich2016b}. Most methods of corpus augmentation are applied to pair monolingual training data with automatic back-translation and then treat them as additional parallel training data. Therefore, we have added comparison experiments. 

We conducted a comparison experiment using 300k sentences as the original data and the remaining 372k sentences as the monolingual data. 

Translation results of comparison experiment are presented in Table \ref{tab:results-mono}. ``+Proposed'' back-translates 508k and 513k from the ``300k+mono'' (672k training data), so that the numbers of sentence pairs are increased from 672k to 2,255k and 2,200k in both directions. 
The proposed method produced higher BLEU scores than the original monolingual method.
These comparison experiments demonstrate that our proposed method can augment the extended data by the other corpus augmentation methods to yield better translation performance. In the future we plan to combine the proposed methods with other augmentation approaches as our results suggest it may be more beneficial than only back-translation. Salient benefits of the proposed method are that it requires no monolingual data and that, without changing the neural network architecture, our method can generate more pseudo-parallel sentences. Moreover, it can be combined with other augmentation methods.



\section{Conclusion}
\label{section:con}
In this paper, we proposed a simple but effective approach to augment the NMT corpus for low-resource language pairs by segmenting long sentences in the corpus, using back-translation, and generating pseudo-parallel sentences pairs. We demonstrated that this approach engenders generation of more pseudo-parallel sentences. Consequently, we obtained higher translation quality for NMT. 
Future studies should include more comparative experiments using other language pairs with different amounts of data.


\begin{thebibliography}{12}
\expandafter\ifx\csname natexlab\endcsname\relax\def\natexlab#1{#1}\fi

\bibitem[{Currey et~al.(2017)Currey, Miceli~Barone, and Heafield}]{Currey2017}
Anna Currey, Antonio~Valerio Miceli~Barone, and Kenneth Heafield. 2017.
\newblock \href {https://doi.org/10.18653/v1/W17-4715} {{Copied Monolingual
  Data Improves Low-Resource Neural Machine Translation}}.
\newblock In \emph{Proceedings of the Second Conference on Machine
  Translation}, pages 148--156. Association for Computational Linguistics.

\bibitem[{Fadaee et~al.(2017)Fadaee, Bisazza, and Monz}]{Fadaee2017}
Marzieh Fadaee, Arianna Bisazza, and Christof Monz. 2017.
\newblock \href {https://doi.org/10.18653/v1/P17-2090} {{Data Augmentation for
  Low-Resource Neural Machine Translation}}.
\newblock In \emph{Proc.\ 55th Annual Meeting of the Assoc.\ for Computational
  Linguistics (Volume 2: Short Papers)}, pages 567--573, Vancouver, Canada.

\bibitem[{Gwinnup et~al.(2017)Gwinnup, Anderson, Erdmann, Young, Kazi, Salesky,
  Thompson, and Taylor}]{Gwin}
Jeremy Gwinnup, Timothy Anderson, Grant Erdmann, Katherine Young, Michaeel
  Kazi, Elizabeth Salesky, Brian Thompson, and Jonathan Taylor. 2017.
\newblock \href {https://doi.org/10.18653/v1/W17-4728} {{The AFRL-MITLL WMT17
  Systems: Old, New, Borrowed, BLEU}}.
\newblock In \emph{Proceedings of the Second Conference on Machine
  Translation}, pages 303--309. Association for Computational Linguistics.

\bibitem[{He et~al.(2016)He, Xia, Qin, Wang, Yu, Liu, and Ma}]{He2016}
Di~He, Yingce Xia, Tao Qin, Liwei Wang, Nenghai Yu, Tie-Yan Liu, and Wei-Ying
  Ma. 2016.
\newblock \href
  {http://papers.nips.cc/paper/6469-dual-learning-for-machine-translation.pdf}
  {Dual learning for machine translation}.
\newblock In \emph{Advances in Neural Information Processing Systems 29}, pages
  820--828. Curran Associates, Inc.

\bibitem[{Klein et~al.(2017)Klein, Kim, Deng, Senellart, and
  Rush}]{OpenNMT:Klein}
Guillaume Klein, Yoon Kim, Yuntian Deng, Jean Senellart, and Alexander~M. Rush.
  2017.
\newblock \href {http://arxiv.org/abs/1701.02810} {{OpenNMT: Open-Source
  Toolkit for Neural Machine Translation}}.
\newblock \emph{CoRR}, abs/1701.02810.

\bibitem[{Koehn and Knowles(2017)}]{KoehnK17}
Philipp Koehn and Rebecca Knowles. 2017.
\newblock \href {http://arxiv.org/abs/1706.03872} {{Six Challenges for Neural
  Machine Translation}}.
\newblock \emph{CoRR}, abs/1706.03872.

\bibitem[{Lample et~al.(2018)Lample, Ott, Conneau, Denoyer, and
  Ranzato}]{Lample}
Guillaume Lample, Myle Ott, Alexis Conneau, Ludovic Denoyer, and Marc'Aurelio
  Ranzato. 2018.
\newblock \href {http://arxiv.org/abs/1804.07755} {{Phrase-Based {\&} Neural
  Unsupervised Machine Translation}}.
\newblock \emph{CoRR}, abs/1804.07755.

\bibitem[{Luong et~al.(2015)Luong, Pham, and Manning}]{Luong2015}
Minh~Thang Luong, Hieu Pham, and Christopher~D. Manning. 2015.
\newblock \href {https://doi.org/10.18653/v1/D15-1166} {{Effective Approaches
  to Attention-based Neural Machine Translation}}.
\newblock In \emph{Proc.\ 2015 Conf.\ on Empirical Methods in Natural Language
  Processing}, pages 1412--1421, Lisbon, Portugal. ACL.

\bibitem[{Nakazawa et~al.(2016)Nakazawa, Yaguchi, Uchimoto, Utiyama, Sumita,
  Kurohashi, and Isahara}]{NAKAZAWA16.621}
Toshiaki Nakazawa, Manabu Yaguchi, Kiyotaka Uchimoto, Masao Utiyama, Eiichiro
  Sumita, Sadao Kurohashi, and Hitoshi Isahara. 2016.
\newblock {ASPEC: Asian Scientific Paper Excerpt Corpus}.
\newblock In \emph{Proceedings of the Tenth International Conference on
  Language Resources and Evaluation (LREC 2016)}, Paris, France.

\bibitem[{Poncelas et~al.(2018)Poncelas, Shterionov, Way, de~Buy~Wenniger, and
  Passban}]{invest}
Alberto Poncelas, Dimitar Shterionov, Andy Way, Gideon~Maillette
  de~Buy~Wenniger, and Peyman Passban. 2018.
\newblock \href {http://arxiv.org/abs/1804.06189} {{Investigating
  Backtranslation in Neural Machine Translation}}.
\newblock \emph{CoRR}, abs/1804.06189.

\bibitem[{Sennrich et~al.(2017)Sennrich, Birch, Currey, Germann, Haddow,
  Heafield, Barone, and Williams}]{Sennrich2017}
Rico Sennrich, Alexandra Birch, Anna Currey, Ulrich Germann, Barry Haddow,
  Kenneth Heafield, Antonio Valerio~Miceli Barone, and Philip Williams. 2017.
\newblock \href {http://arxiv.org/abs/1708.00726} {{The University of
  Edinburgh's Neural {MT} Systems for {WMT17}}}.
\newblock \emph{CoRR}, abs/1708.00726.

\bibitem[{Sennrich et~al.(2015)Sennrich, Haddow, and Birch}]{Sennrich2016b}
Rico Sennrich, Barry Haddow, and Alexandra Birch. 2015.
\newblock \href {http://arxiv.org/abs/1511.06709} {{Improving Neural Machine
  Translation Models with Monolingual Data}}.
\newblock \emph{CoRR}, abs/1511.06709.

\end{thebibliography}

\end{document}